\documentclass[tablecaption=bottom,wcp]{jmlr} 
\usepackage[utf8]{inputenc}

\usepackage{booktabs}

\theorembodyfont{\upshape}
\theoremheaderfont{\scshape}
\theorempostheader{:}
\theoremsep{\newline}

\usepackage{caption}
\usepackage{subcaption}
\usepackage{listings, chngcntr, float, amsmath, amssymb, cmap, graphicx, xcolor, hyperref}
\usepackage{amsmath, multirow}

\usepackage{color}
\usepackage{grffile}
\usepackage{array,tabulary}
\usepackage{wrapfig}
\graphicspath{{media/}}

\usepackage{mathtools,mathptmx}
\mathtoolsset{showonlyrefs=true}

\usepackage[mathcal]{euscript}
\usepackage{amsfonts}
\usepackage{tikz}
\usetikzlibrary{arrows.meta,shapes.misc}

\jmlrvolume{60}
\jmlryear{2017}
\jmlrworkshop{Conformal and Probabilistic Prediction and Applications}
\jmlrproceedings{PMLR}{Proceedings of Machine Learning Research}

\title[Conformal k-NN Anomaly Detector for Univariate Data Streams]{Conformal $k$-NN Anomaly Detector for Univariate Data Streams}

\newcommand*{\mathcolor}{}
\def\mathcolor#1#{\mathcoloraux{#1}}
\newcommand*{\mathcoloraux}[3]{%
  \protect\leavevmode
  \begingroup
    \color#1{#2}#3%
  \endgroup
}

\renewcommand{\vec}{\mathbf} 

\author{\Name{Vladislav Ishimtsev} \Email{vladislav.ishimtsev@skolkovotech.ru}\\
\addr Skolkovo Institute of Science and Technology,  Skolkovo, Moscow Region, Russia\\
\addr Institute for Information Transmission Problems, Moscow, Russia
\AND
\Name{Alexander Bernstein} \Email{a.bernstein@skoltech.ru}\\
\addr Skolkovo Institute of Science and Technology,  Skolkovo, Moscow Region, Russia\\
\addr Institute for Information Transmission Problems, Moscow, Russia
\AND
\Name{Evgeny Burnaev}\Email{e.burnaev@skoltech.ru}\\
\addr Skolkovo Institute of Science and Technology,  Skolkovo, Moscow Region, Russia\\
\addr Institute for Information Transmission Problems, Moscow, Russia
\AND
\Name{Ivan Nazarov} \Email{ivan.nazarov@skolkovotech.ru}\\
\addr Skolkovo Institute of Science and Technology,  Skolkovo, Moscow Region, Russia\\
\addr Institute for Information Transmission Problems, Moscow, Russia
}

\editors{Alex Gammerman, Vladimir Vovk, Zhiyuan Luo, and Harris Papadopoulos}

\begin{document}

\maketitle
\medskip

\begin{abstract}
Anomalies in time-series data give essential and often actionable information in
many applications. In this paper we consider a model-free anomaly detection method
for univariate time-series which adapts to non-stationarity in the data stream and
provides probabilistic abnormality scores based on the conformal prediction paradigm.
Despite its simplicity the method performs on par with complex prediction-based
models on the Numenta Anomaly Detection benchmark and the Yahoo! S5 dataset.
\end{abstract}


\section{Introduction} 
\label{sec:introduction}

Anomaly detection in time-series data has important applications in many practical
fields \citep{kejariwal2015}, such as monitoring of aircraft's cooling systems in
aerospace industry \citep{alestraetal2014}, detection of unusual symptoms in healthcare,
monitoring of software-intensive systems \citep{artemovburnaev16}, of suspicious
trading activity by regulators or high frequency dynamic portfolio management in
finance, etc.

General anomaly detection methods can be broadly categorized in five families, \citep{pimenteletal2014},
each approaching the problem from a different angle: probabilistic, distance-based,
prediction-based, domain-based, and information-theoretic techniques. The common
feature of all families is the reliance on a negative definition of abnormality:
``abnormal'' is something which is not ``normal'', i.e. a substantial deviation from
a typical set of patterns.

Prediction-based anomaly detection techniques rely on an internal regression model
of the data: for each test example the discrepancy between the observed and the
prediction, i.e. the reconstruction error, is used to decide its abnormality. For
example, neural networks are used in this manner in \citep{augusteijnfolkert2002}
and \citep{hawkinsetal2002,williamsetal2002}, whereas in \citep{chandolaetal2009}
the predictions are based on a comprehensive description of the variability of the
input data. Other reconstruction methods include dimensionality reduction \citep{jolliffe2014},
linear and kernel Principal Component Analysis \citep{duttaetal2007,shyuetal2003,hoffmann2007,scholkopfetal1998}.

Anomaly detection in time-series analysis is complicated by high noise and the fact
that the assumptions of classical change point models are usually violated by either
non-stationarity or quasi-periodicity of the time-series \citep{artemovetal2015,artemovburnaev16},
or long-range dependence \citep{artemovburnaev2015a}. Classical methods require strong
pre- and post- change point distributional assumptions, when in reality change-points
might exhibit clustering, or be starkly contrasting in nature between one another.
Thus, the usual approach of detecting anomalies against a fixed model, e.g. the classical
models \citep{burnaev2009,burnaevetal2009}, is unsubstantiated. This has compelled
practitioners to consider specialized methods for anomaly model selection \citep{burnaevetal2015a},
construction of ensembles of anomaly detectors \citep{artemovburnaev2015b}, and explicit
rebalancing of the normal and abnormal classes \citep{burnaevetal2015b}, among others.

Time-series anomaly detection techniques include, among others, spatiotemporal self
organising maps \citep{barretoetal2009}, kurtosis-optimising projections of a VARMA
model used as features for outlier detection algorithm based on the CUSUM \citep{galeanoetal2006},
Multidimensional Probability Evolution method to identify regions of the state space
frequently visited during normal behaviour \citep{leeroberts2008}, tracking the empirical
outlier fraction of the one-class SVM on sliding data slices \citep{gardneretal2006},
or applying one-class SVM to centred time-series embedded into a phase space by
a sliding window \citep{maperkins2003}. The main drawback of these approaches is
that they use explicit data models, which require parameter estimation and model
selection.

Distance-based anomaly detection methods perform a task similar to that of estimating
the pdf of data and do not require prior model assumptions. They rely on a metric,
usually Euclidean of Mahalanobis, to quantify the degree of dissimilarity between
examples and to derive either a distance-based or a local density score in order to
assess abnormality. Such methods posit that the normal observations are well embedded
within their metric neighbourhood, whereas outliers are not.

Despite being model-free, distance-based methods do not provide a natural probabilistic
measure, which conveys detector's degree of confidence in abnormality of an observation.
Indeed, there do exist distance-based methods, for example LoOP, \citep{kriegeletal2009},
which output this kind of score, but typically they rely on quite limiting distributional
assumptions. Such assumptions can potentially be avoided by using conformal prediction
methods, \citep{shaferetal2008}. For instance, conformal prediction allows efficient
construction of non-parametric confidence intervals \citep{nazarov2016}.

This paper outlines an anomaly detection method in univariate time-series, which
attempts to adapt to non-stationarity by computing ``deferred'' scores and uses
conformal prediction to construct a non-parametric probabilty measure, which efficiently
quantifies the degree of confidence in abnormality of new observations. We also
provide technical details on boosting the performance of the final anomaly detector,
e.g. signal pruning. The extensive comparison on Yahoo! S5 and Numenta benchmark
datasets revealed that the proposed method performs on par with complex prediction-based
detectors. The proposed method is among the top 3 winning solutions of the 2016 Numenta
Anomaly Detection Competition, see \citep{numenta2016}.

In section~\ref{sec:conformal_anomaly_detection} we review general non-parametric
techniques for assigning confidence scores to anomaly detectors. In sec.~\ref{sec:anomaly_detection_in_uts}
we propose a conformal detector for univariate time-series based on $k$-NN ($k$
Nearest Neighbours) and time-delay embedding, which attempts to tackle quasi-periodicity
and non-stationarity issues. In section~\ref{sec:anomaly_detection_benchmark} we
provide details on the comparison methodology and the Numenta Anomaly Detection benchmark,
and in section~\ref{sec:benchmark_results} we compare the performance of the proposed
method.


\section{Conformal Anomaly Detection} 
\label{sec:conformal_anomaly_detection}

Conformal Anomaly Detection (CAD), \citep{laxhammar2014}, is a distribution-free
procedure, which assigns a probability-like confidence measure to predictions of
an arbitrary anomaly detection method. CAD uses the scoring output of the detector
$A(X_{:t}, \vec{x}_{t+1})$ as a measure of non-conformity (Non-Conformity Measure,
NCM), which quantifies how much different a test object $\vec{x}_{t+1} \in \mathcal{X}$
is with respect to the \textit{reference} sample $X_{:t}=(\vec{x}_s)_{s=1}^t \in \mathcal{X}$.
Typical examples of NCMs are prediction error magnitude for a regression model,
reconstruction error for dimensionality reduction methods, average distance to the
$k$ nearest neighbours, etc. The NCM may have intrinsic randomness independent of
the data, \citep{vovk2013}. For a sequence of observations $\vec{x}_t \in \mathcal{X}$,
$t = 1,2,\ldots$, at each $t\geq 1$ CAD computes the scores
\begin{equation} \label{eq:cad_scores}
  \alpha_s^t
    = A(X_{:t}^{-s}, \vec{x}_s)
    \,,\, s = 1,\ldots,t
    \,,
\end{equation}
where $X_{:t}^{-s}$ is the sample $X_{:t}$ without the $s$-th observation. The
confidence that $\vec{x}_t$ is anomalous relative to the \textit{reference}
sample $X_{:(t-1)}$ is one minus the empirical $p$-value of its non-conformity
score $\alpha_t^t$ in \eqref{eq:cad_scores}:
\begin{equation} \label{eq:cad_p_value}
  p(\vec{x}_t, X_{:(t-1)}, A)
    = \frac1t \Bigl|
          \{s=1,\ldots,t \,:\, \alpha_s^t \geq \alpha_t^t\}
        \Bigr|
    \,.
    \tag{CPv}
\end{equation}
Basically, the more abnormal $\vec{x}_t$ is the lower its $p$-value is, since
anomalies, in general, poorly conform to the previously observed reference sample
$X_{:(t-1)}$.

In \citep{shaferetal2008} it was shown that online conformal prediction, and by
extension CAD, offers conservative coverage guarantees in online learning setting.
Indeed, when iid sequence $\vec{x}_t \sim D$ is fed into the conformal anomaly
detector one observation at a time, then for any NCM $A$ and all $t\geq 1$
\begin{equation} \label{eq:cad_coverage}
  \mathbb{P}_{X \sim D^t} \bigl(
        p(\vec{x}_t, X^{-t}, A) < \epsilon
      \bigr)
    \leq \epsilon
    \,,\, X=(\vec{x}_s)_{s=1}^t
    \,.
\end{equation}
Intuitively, \eqref{eq:cad_p_value} is the empirical CDF, obtained on a sample
$(A(X^{-s}, \vec{x}_s))_{s=1}^t$, evaluated at a random point $A(X^{-t}, \vec{x}_t)$
with the sample $X$ drawn from an exchangeable distribution $D^t$. This means that
the distribution of the $p$-value itself is asymptotically uniform. The NCM, used
in \eqref{eq:cad_p_value}, affects the tightness of the guarantee \eqref{eq:cad_coverage}
and the volume of computations.

At any $t\geq1$ in \eqref{eq:cad_scores} CAD requires $t$ evaluations of $A$ with
different samples $X_{:t}^{-s}$, which is potentially computationally heavy.
\citep{laxhammaretal2015} proposed the Inductive Conformal Anomaly Detection (ICAD)
which uses a fixed proper training sample of size $n$ as the reference in the non-%
conformity scores. If the sequence $(\vec{x}_t)_{t\geq1}$ is relabelled so that it
starts at $1-n$ instead of $1$, then for each $t\geq1$ the ICAD uses the following
setup:
\begin{equation*}
      \underbrace{
          \vec{x}_{-n+1}
        , \ldots
        , \vec{x}_0
      }_{\tilde{X} \text{ proper training}}
    , \overbrace{\mathcolor{blue}{
          \vec{x}_1
        , \vec{x}_2
        , \ldots
        , \vec{x}_{t-1}
      }}^{\text{calibration}}
    , \underbracket{\mathcolor{red}{
        \vec{x}_t
      }}_{\text{test}}
    , \ldots \,.
\end{equation*}
The conformal $p$-value of a test observation $\vec{x}_t$ is computed using \eqref{eq:cad_p_value}
on the modified scores:
\begin{equation} \label{eq:icad_scores}
  \alpha_s^t
    = A\bigl(\tilde{X}, \vec{x}_s\bigr)
    \,,\, s = 1,\ldots,t
    \,,\, \tilde{X} = (\vec{x}_{-n+1},\ldots,\vec{x}_0)
    \,.
\end{equation}
The ICAD is identical to CAD over the sequence $(\vec{x}_t)_{t\geq n+1}$ (relabelled
to start at $1$) with a non-conformity measure $\bar{A}$, which always \textit{ignores}
the supplied \textit{reference} sample and uses the proper training sample $\tilde{X}$
instead. Therefore the ICAD has similar coverage guarantee as \eqref{eq:cad_coverage}
with scores given by \eqref{eq:icad_scores}.

By trading the deterministic guarantee \eqref{eq:cad_coverage} for a PAC guarantee
it is possible to make the ICAD use a fixed-size calibration set. The resulting
``sliding'' ICAD fixes the size of the calibration sample to $m$ and forces
it to move along the sequence $(\vec{x}_t)_{t\geq1}$, i.e.
\begin{equation*}
      \underbrace{
          \vec{x}_{-n+1}
        , \ldots
        , \vec{x}_0
      }_{\tilde{X} \text{ training}}
    , \, \ldots \,
    , \overbrace{\mathcolor{blue}{
          \vec{x}_{t-m}
        , \ldots
        , \vec{x}_{t-1}
      }}^{\text{calibration}}
    , \underbracket{\mathcolor{red}{
        \vec{x}_t
      }}_{\text{test}}
    , \ldots
    \,.
\end{equation*}
The conformal $p$-value uses a subsample of the non-conformity scores \eqref{eq:icad_scores}:
\begin{equation} \label{eq:sliding_icad_p_value}
  p(\vec{x}_t, X_{:(t-1)}, A)
    = \frac1{m+1} \Bigl|
          \{i=0,\ldots,m\,:\, \alpha_{t-i}^t \geq \alpha_t^t\}
        \Bigr|
    \,.
    \tag{$\text{CPv}_m$}
\end{equation}
The guarantee for ICAD is a corollary to proposition~(2) in \citep{vovk2013}. In fact,
the exchangeability of $(\vec{x}_t)_{t\geq1}$ further implies a similar PAC-type
validity result for the sliding ICAD, which states that for any $\delta, \epsilon\in(0,1)$
for any fixed proper training set $\tilde{X}$ and data distribution $D$ it is true that
\begin{equation} \label{eq:sliding_icad_validity}
  \mathbb{P}_{\vec{x} \sim D} \bigl(
        p(\vec{x}, X, \bar{A}) < \epsilon
      \bigr)
    \leq \epsilon + \sqrt{\frac{\log\frac1{\delta}}{2m}}
    \,,
\end{equation}
with probability at least $1-\delta$ over draws of $X \sim D^m$, and $\bar{A}$ is
the NCM $\vec{x}\mapsto A(\tilde{X}, \vec{x})$, which uses $\tilde{X}$ as the
\textit{reference} sample.


\section{Anomaly Detection in Univariate Time Series} 
\label{sec:anomaly_detection_in_uts}

In this section we outline the building blocks of the proposed model-free detection
method which produces conformal confidence scores for its predictions. The conformal
scores are computed using an adaptation of the ICAD to the case of potentially
non-stationary and quasi-periodic time-series.

Consider a univariate time-series $X = (x_t)_{t\geq1} \in \real$. The first step of
the proposed procedure is to embed $X$ into an $l$-dimensional space, via a sliding
historical window:
\begin{equation} \label{eq:time_delay_embed}
    \ldots
  , x_{t-l-1}
  , \rlap{$\overbracket{
      \phantom{
          x_{t-l}
        , x_{t-l+1}
        , \ldots
        , x_{t-1}
      }}^{\vec{x}_{t-1}}$
    }
    \mathcolor{red}{x_{t-l}}
  , \underbracket{
      \mathcolor{blue}{
          x_{t-l+1}
        , \ldots
        , x_{t-1}
      }
      , \mathcolor{red}{x_t}
    }_{\vec{x}_t}
  , x_{t+1}
  , \ldots
  \,. \tag{T-D}
\end{equation}
In other words, $\vec{x}_t\in \real^l$ is $l$ most recent observations of $x_s$,
$s=t-l+1,\ldots,t$. This embedding requires a ``burn-in'' period of $l$ observations
to accumulate at least one full window, unless padding is used.

This embedding of $X$ permits the use of multivariate distance-based anomaly detection
techniques. Distance-based anomaly detection uses a distance $d$ on the input space
$\mathcal{X}$ to quantify the degree of dissimilarity between observations. Such
methods posit that the normal observations are generally closer to their neighbours,
as opposed to outlying examples which typically lie farther. If the space $\mathcal{X}$
is $\real^{d\times 1}$ then, the most commonly used distance is the Mahalanobis metric,
which takes into account the general shape of the sample and correlations of the
data. In the following the distance, induced by the sample $\mathcal{S}=(\vec{x}_i)_{i=1}^n$,
is $d(\vec{x}, \vec{y}) = \sqrt{(\vec{x}-\vec{y})' {\hat{\Sigma}}^{-1} (\vec{x}-\vec{y})}$,
where $\hat{\Sigma}$ is an estimate of the covariance matrix on $\mathcal{S}$.

The $k$-NN anomaly detector assigns the abnormality score to some observation $\vec{x}%
\in \mathcal{X}$ based on the neighbourhood proximity measured by the average distance
to the $k$ nearest neighbours:
\begin{equation} \label{eq:k_nn_scorer}
  \text{NN}(\vec{x}; k, \mathcal{S})
    = \frac1{|N_k(\vec{x})|} \sum_{\vec{y} \in N_k(\vec{x})} d(\vec{x},\vec{y})
    \,,
\end{equation}
where $N_k(\vec{x})$ are the $k$ nearest neighbours of $\vec{x}$ within $\mathcal{S}$
excluding itself. The detector labels as anomalous any observation with the score
exceeding some calibrated threshold. The main drawbacks are high sensitivity to $k$
and poor interpretability of the score $\text{NN}(\vec{x}; k)$, due to missing natural
data-independent scale. Various modifications of this detector are discussed in
\citep{ramaswamyetal2000,angiullietal2002,bayetal2003,hautamakietal2004} and \citep{zhangwang2006}.

Alternatively, it is also possible to use density-based detection methods. For example
the schemes proposed in \citep{breunigetal2000,kriegeletal2009} are based on $k$-NN,
but introduce the concept of \emph{local data density}, a score that is inversely
related to a distance-based characteristic of a point within its local neighbourhood.
Similarly to the $k$-NN detector, these methods lack a natural scale for the abnormality
score. Modifications of this algorithm are discussed in \citep{jinetal2006} and
\citep{papadimitriouetal2003}.

The combination of the embedding \eqref{eq:time_delay_embed} and the scoring function
\eqref{eq:k_nn_scorer} produces a non-conformity measure $A$ for conformal procedures
in sec.~\ref{sec:conformal_anomaly_detection}. The most suitable procedure is the
sliding ICAD, since CAD and the online ICAD are heavier in terms of runtime complexity
(tab.~\ref{tab:method_cplx}). However, the sliding ICAD uses a fixed proper training
sample for \textit{reference}, which may not reflect potential non-stationarity.
Therefore we propose a modification called the Lazy Drifting Conformal Detector (LDCD)
which adapts to normal regime non-stationarity, such as quasi-periodic or seasonal
patterns. The LDCD procedure is conceptually similar to the sliding ICAD, and thus
is expected to provide similar validity guarantees at least in the true iid case.
The main challenge is to assess the effects of the calibration scores within the
same window being computed on different windows of the training stream.

For the observed sequence $(\vec{x}_t)_{t\geq1}$, the LDCD maintains two fixed-size
separate samples at each moment $t\geq n+m$: the training set $\mathcal{T}_t = (\vec{x}_{t-N+i})_{i=0}^{n-1}$
of size $n$ ($N=m+n$) and the calibration \textbf{queue} $\mathcal{A}_t$ of size $m$.
The sample $\mathcal{T}_t$ is used as the \textit{reference} sample for conformal
scoring as in \eqref{eq:cad_scores}. The calibration \textbf{queue} $\mathcal{A}_t$
keeps $m$ most recent non-conformity scores given by $\alpha_s = A(\mathcal{T}_s, \vec{x}_s)$
for $s=t-m,\ldots,t-1$. At each $t\geq n+m$ the samples $\mathcal{A}_t$ and
$\mathcal{T}_t$ look as follows:
\begin{equation*}
  \begin{aligned}
    \text{data: }
      & \ldots
      , \overbrace{
          \vec{x}_{t-m-n}
        , \ldots
        , \vec{x}_{t-m-1}
        }^{\mathcal{T}_t \text{ training}}
      , & \vec{x}_{t-m}
      , \ldots
      , \vec{x}_{t-1}
      , &\,& \overbracket{\mathcolor{red}{\vec{x}_t}}^{\text{test}}
      , \ldots \\
    \text{scores: }
      & \ldots
      , \alpha_{t-m-n}
      , \ldots
      , \alpha_{t-m-1}
      , & \underbrace{\mathcolor{blue}{
          \alpha_{t-m}
        , \ldots
        , \alpha_{t-1}
        }}_{\mathcal{A}_t \text{ calibration}}
      , &\,& \overbracket{\mathcolor{red}{\alpha_t}}^{\text{test}}
      , \ldots
  \end{aligned}
\end{equation*}
The procedure uses the current test observation $\vec{x}_t$ to compute the non-conformity
score $\alpha_t$ used to obtain the $p$-value similarly to \eqref{eq:sliding_icad_p_value},
but with respect to scores in the calibration queue $\mathcal{A}_t$. At the end of
step $t$ the calibration queue is updated by pushing $\alpha_t$ into $\mathcal{A}_t$
and evicting $\alpha_{t-m}$.

The final conformal $k$-NN anomaly detector is defined by the following procedure:
\begin{enumerate}
  \item the time-series $(x_t)_{t\geq1}$ is embedded into $\real^l$ using
  \eqref{eq:time_delay_embed} to get the sequence $(\vec{x}_{t+l-1})_{t\geq 1}$;
  \item the LDCD uses $k$-NN average distance \eqref{eq:k_nn_scorer} for scoring
  $(\vec{x}_t)_{t\geq 1}$.
\end{enumerate}
The proper training sample $\mathcal{T}_t$ for $t=n+m+1$ is initialized to the first
$n$ observations of $\vec{x}_t$, and the calibration queue $\mathcal{A}_t$ and is
populated with the scores $\alpha_{n+s} = \text{NN}(\vec{x}_{n+s}; k, \mathcal{T}_{n+m+1})$
for $s=1,\ldots,m$.



\section{Anomaly Detection Benchmark} 
\label{sec:anomaly_detection_benchmark}

The Numenta Anomaly Benchmark (NAB), \citep{lavinetal2015}, is a corpus of datasets
and a rigorous performance scoring methodology for evaluating algorithms for online
anomaly detection. The goal of NAB is to provide a controlled and repeatable environment
for testing anomaly detectors on data streams. The scoring methodology permits only
automatic online adjustment of hyperparameters to each dataset in the corpus during
testing. In this study we supplement the dataset corpus with additional data (sec.~%
\ref{sub:datasets}), but employ the default NAB scoring methodology (sec.~%
\ref{sub:performance_scoring}).

\subsection{Datasets} 
\label{sub:datasets}

\begin{figure}[b]
  \begin{minipage}[ht]{0.48\linewidth}
    \centering
    Yahoo! Corpus
    \begin{minipage}[ht]{\linewidth}
      \center{\includegraphics[width=\linewidth]{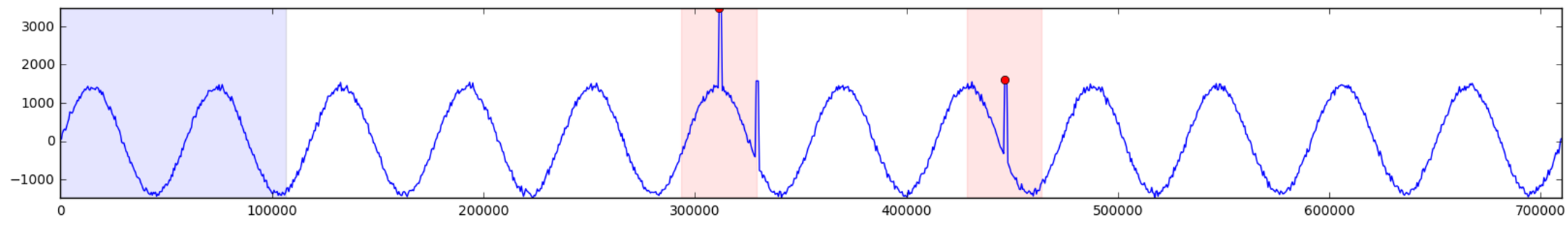} \\ }
    \end{minipage} 
    \vfill
    \begin{minipage}[ht]{\linewidth}
      \center{\includegraphics[width=\linewidth]{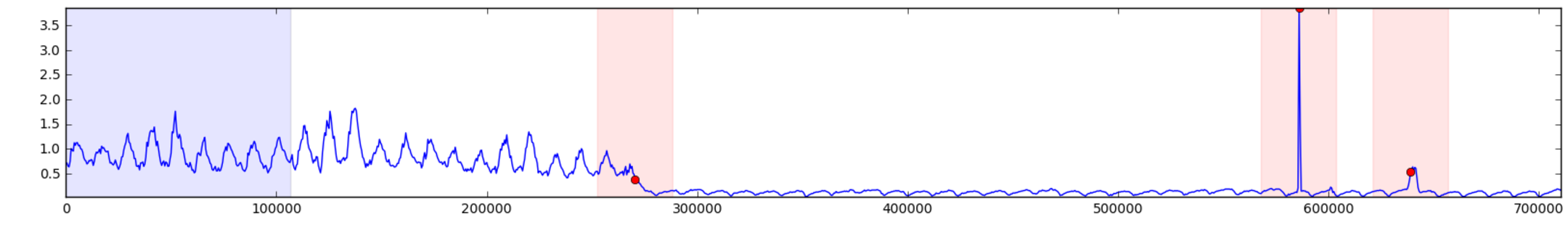} \\ }
    \end{minipage} 
    \vfill
    \begin{minipage}[ht]{\linewidth}
      \center{\includegraphics[width=\linewidth]{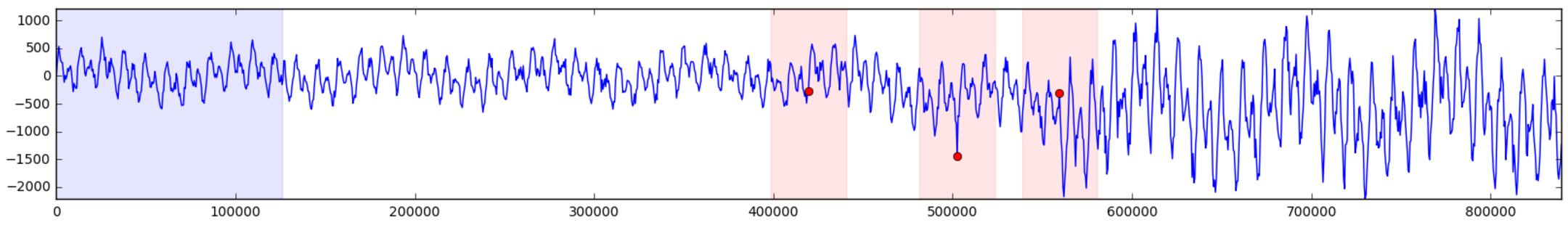} \\}
    \end{minipage}
  \end{minipage}
  \hfill
  \begin{minipage}[ht]{0.48\linewidth}
    \centering
    NAB Corpus
    \begin{minipage}[ht]{\linewidth}
      \center{\includegraphics[width=\linewidth]{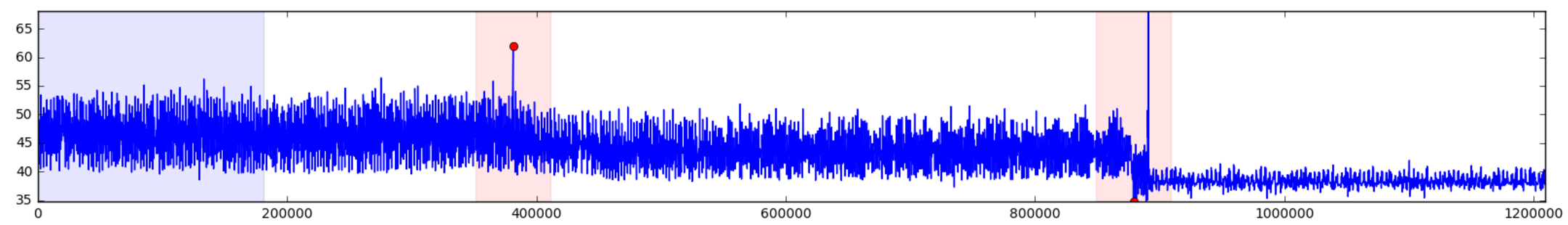} \\ }
    \end{minipage} 
    \vfill
    \begin{minipage}[ht]{\linewidth}
      \center{\includegraphics[width=\linewidth]{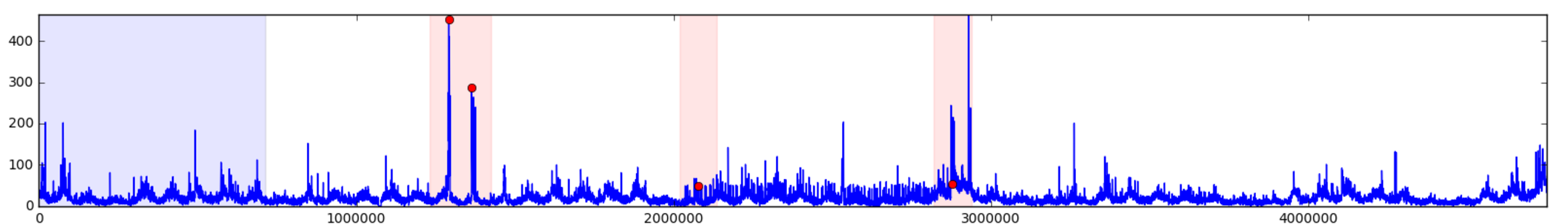} \\ }
    \end{minipage} 
    \vfill
    \begin{minipage}[ht]{\linewidth}
      \center{\includegraphics[width=\linewidth]{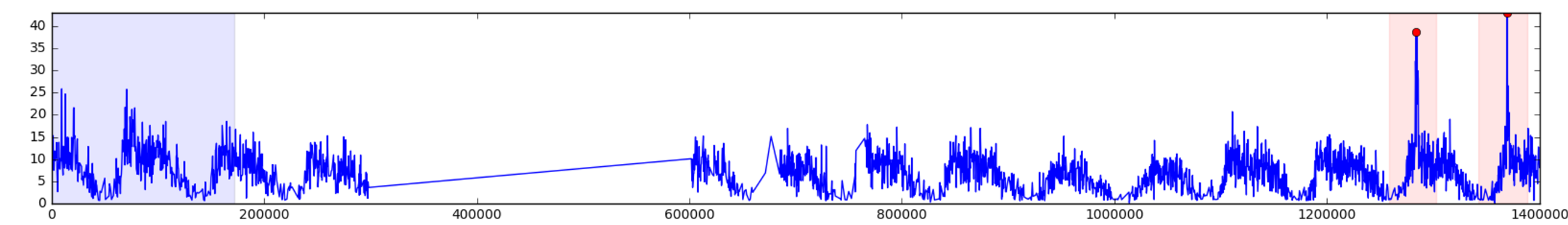} \\ }
    \end{minipage}
  \end{minipage}  
  \caption{Examples of time-series data from Yahoo! and NAB corpora. The
  {\color{red} red} shaded regions represent the anomaly windows centered
  at anomalies. The {\color{blue} blue} region marks the data which the
  benchmark offers for initial parameter estimation and hyperparameter
  tuning.}
  \label{fig:example_datasets}
\end{figure}

The NAB corpus contains $58$ real-world and artificial time-series with $1000$-$22000$
observations per series. The real data ranges from network traffic and CPU utilization
in cloud services to sensors on industrial machines and social media activity. The
dataset is labelled manually and collaboratively according to strict and detailed
guidelines established by Numenta. Examples of time-series are provided in fig.~%
\ref{fig:example_datasets}.

We supplement the NAB corpus with Yahoo! S5 dataset, \citep{yahoos5}, which was collected
to benchmark detectors on various kinds of anomalies including outliers and change-%
points. The corpus contains $367$ tagged real and synthetic time-series, divided
into $4$ subsets. The first group contains real production metrics of various Yahoo!
services, and the other 3 -- synthetic time-series with varying trend, noise and
seasonality, which include either only outliers, or both outliers and change-points.
We keep all univariate time-series from first two groups for benchmarking. Statistics
of the datasets in each corpus are given in tab.~\ref{tab:corpora_stat}.

\begin{table}[t]
  \centering
  \begin{tabular}{l|lc|cccc|cccc}
    \toprule
        \multirow{2}{*}{Corpus}
      & \multirow{2}{*}{Type}
      & \multirow{2}{*}{datasets}
      & \multicolumn{3}{c}{Observations}
      & \multirow{2}{*}{Total}
      & \multicolumn{3}{c}{Anomalies}
      & \multirow{2}{*}{Total}
    \\ 
      & & & Min & Mean & Max
        & & Min & Mean & Max
      &
    \\ 
    \midrule
    \multirow{3}{*}{Yahoo!}
      & Synthetic
        &  33 & 1421 & 1594 & 1680 &  52591 & 1 & 4.03 & 8 & 133 \\
      & Real
        &  67 &  741 & 1415 & 1461 &  94866 & 0 & 2.13 & 5 & 143 \\
      & Total
        & 100 &  741 & 1475 & 1680 & 147457 & 0 & 2.76 & 8 & 276 \\
    \midrule
    \multirow{3}{*}{NAB}
      & Synthetic
        & 11 & 4032 & 4032 &  4032 &  44352 & 0 & 0.55 & 1 &   6 \\
      & Real
        & 47 & 1127 & 6834 & 22695 & 321206 & 0 & 2.43 & 5 & 114 \\
      & Total
        & 58 & 1127 & 6302 & 22695 & 365558 & 0 & 2.07 & 5 & 120 \\
     \bottomrule
  \end{tabular}
  \caption{Description of the NAB and Yahoo! S5 corpora.}
  \label{tab:corpora_stat}
\end{table}


\subsection{Performance scoring} 
\label{sub:performance_scoring}

\begin{figure}[t]
  \centering
  \begin{tikzpicture}[scale=0.75]
    \begin{scope}[scale=2.0]
  \clip(-3.25,-.5) rectangle (4.5,3.5);
  \tikzset{cross/.style={cross out, draw=black, minimum size=2*(#1-\pgflinewidth), inner sep=0pt, outer sep=0pt}, cross/.default={1pt}}

  \draw[black,-{>[scale=2.5,length=1.5,width=.75]}]
    (-3,0) -- (4,0) node[below] { \texttt{time}};
  \draw[black,dotted] (0,0) -- (0,3.0) {};
  \draw[black] (0,0) node[cross=2pt] {};
  \draw[black] (0,0) node[below] { \texttt{event}};

  \draw[black,dotted] (-.75,0) -- (-.75,3.25) {};
  \draw[black,fill=white] (-.75,0) circle(.75pt) {};
  \draw[black,dotted] (0.75,0) -- (0.75,3.25) {};
  \draw[black,fill=white] (0.75,0) circle(.75pt) {};

  \draw[black,{<[scale=2.5,length=.75,width=.5]}-{>[scale=2.5,length=.75,width=.5]}]
    (-.75,3.0) -- (0.75,3.0) node[midway,above] { \texttt{true positive}};

  \draw[black] (-2,0.75) node[left] { $-1$};
  \draw[blue,solid,thick] (-2,0.75) -- (-.75,0.75);
  \draw[blue,fill=white] (-.75,0.75) circle(0.75pt);


  \draw[blue,fill=blue] (-.75,2.75) circle(0.75pt);
  \draw[black] (-.75,2.75) node[left] { $+1$};
  \draw[blue,solid,thick,domain=-.75:1.5,smooth,variable=\x,samples=120]
    plot ({\x},{0.75 + 2 / (1 + exp(10*(\x - 0.75)))});

  \draw[blue,fill=blue] (0.75,1.75) circle(0.75pt);
  \draw[black] (0.75,1.75) node[right] { $0$};

  \draw[blue,solid,thick] (1.5,0.75) -- (3,0.75);
  \draw[black,dashed] (3,0.75) node[right] { $-1$};


  \draw[black!10!red,fill=black!10!red] (-1.75,0.75) circle(1.25pt);
  \draw[black!10!red,fill=black!10!red] (2.5,0.75) circle(1.25pt);
  \draw[black!10!red,fill=black!10!red]
    (.9,{0.75 + 2 / (1 + exp(10*(.9 -.75)))}) circle(1.25pt);

  \def\xcoord{-.5,-.25,0.3,0.6}
  \foreach[count=\i] \xx in \xcoord {
    \def\yy{0.75 + 2 / (1 + exp(10*(\xx -.75)))};

    \coordinate (pt\i) at ({\xx},{\yy});
    \ifnum \i > 1
      \draw[black!60!green,fill=white] (pt\i) circle(1.25pt);
    \else
      \draw[black!60!green,fill=black!60!green] (pt\i) circle(1.25pt);
    \fi
  }

  \draw[black!60!green,fill=black!60!green] (2.0,2.5) circle(1.25pt);
  \draw[black!10!red,fill=black!10!red] (1.85,2.5) circle(1.25pt);
  \draw[black!60!green,fill=white] (2.0,2.3) circle(1.25pt);
  \draw (2.0,2.5) node[right] {\scriptsize \texttt{counted}};
  \draw (2.0,2.3) node[right] {\scriptsize \texttt{ignored}};

\end{scope}
  \end{tikzpicture}
  \caption{Score weighting in NAB: \textbf{all} detections outside the window are
  {\color{black!10!red} \textbf{false alarms}}, whereas only \textbf{the earliest}
  detection within the window is a {\color{black!60!green} \textbf{true positive}},
  and later detections are ignored.
  }
  \label{fig:nab_scoring}
\end{figure}
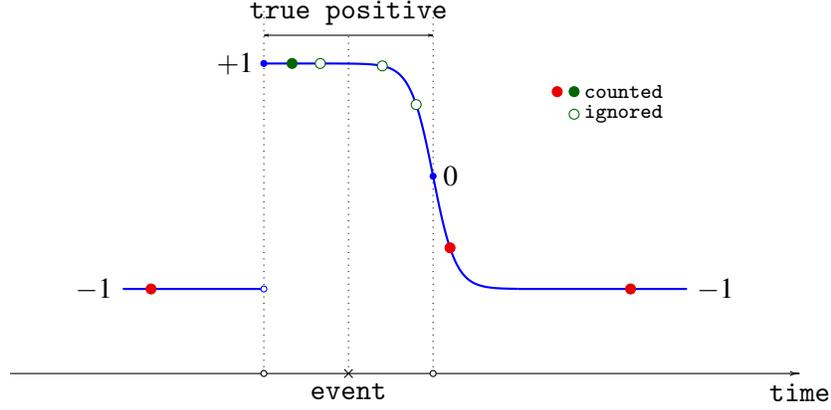

Typical metrics, such as precision and recall, are poorly suited for anomaly detection,
since they do not incorporate time. The Numenta benchmark proposes a scoring methodology,
which favours timely true detections, softly penalizes tardy detections, and harshly
punishes false alarms. The scheme uses anomaly windows around each event to categorize
detections into true and false positives, and employs sigmoid function to assign
weights depending on the relative time of the detection. Penalty for missed anomalies
and rewards for timely detections is schematically shown in fig.~\ref{fig:nab_scoring}.

The crucial feature of scoring is that all false positives decrease the overall score,
whereas only the earliest true positive detection within each window results in a
positive contribution. The number of false negatives is the number of anomaly windows
in the time-series, with no true positive detections. True negatives are not used in
scoring.
\begin{table}[b]
    \centering
    \begin{tabular}{lcccc}
      \toprule
        Metric        & $A_{TP}$ & $A_{FP}$ & $A_{TN}$ & $A_{FN}$ \\
      \midrule
        Standard      & 1.0      & -0.11    & 1.0      & -1.0     \\
        LowFP         & 1.0      & -0.22    & 1.0      & -1.0     \\
        LowFN         & 1.0      & -0.11    & 1.0      & -2.0     \\
      \bottomrule
    \end{tabular}
    \caption{The detection rewards of the default application profiles in the
    benchmark.}
    \label{tab:nad_app_costs}
\end{table}

The relative costs of true positives (TP), false positives (FP) and false negatives
(FN) vary between applications. In NAB this domain specificity is captured by the
\textit{application profile}, which multiplicatively adjusts the score contributions
of TP, FP, and FN detections. NAB includes three prototypical application profiles,
tab.~\ref{tab:nad_app_costs}. The ``Standard'' application profile mimics symmetric
costs of misdetections, while the ``low FP'' and ``low FN'' profiles penalize either
overly optimistic or conservative detectors, respectively. For the anomaly window
of size $\approx 10\%$ of the span of the time-series, the standard profile assigns
relative weights so that random detections made $10\%$ of the time get on average
a zero final score, \citep{lavinetal2015}.

If $X$ is the time-series with labelled anomalies, then the NAB score for a given
detector and application profile is computed as follows. Each detection is matched
to the anomaly window with the nearest right end after it. If $\tau$ is the relative
position of a detection with respect to the right end of the anomaly window of width
$W$, then the score of this detection is
\begin{equation*}
  \sigma(\tau)
    = \begin{cases}
      A_{FP},
          &\text{ if } \tau < -W \,; \\
      (A_{TP} - A_{FP}) \bigl(1 + e^{5 \tau}\bigr)^{-1} + A_{FP},
          &\text{ otherwise } \,.
    \end{cases}
\end{equation*}
The overall performance of the detector over $X$ under profile $A$ is the sum of
the weighted rewards from individual detections and the impact of missing windows.
It is given by
\begin{equation*}
  S_\mathtt{det}^A(X)
    = \sum_{d\in D_\mathtt{det}(X)} \sigma(\tau_d)
      + A_{FN} f_\mathtt{det} \,,
\end{equation*}
where $D_\mathtt{det}(X)$ is the set of all alarms fired by the detector on the
stream $X$, $\tau_d$ is the relative position of a detection $d\in D_\mathtt{det}(X)$,
and $f_\mathtt{det}$ is the number of anomaly windows which cover no detections
at all. The raw benchmark score $S_\mathtt{det}^A$ of the detector is the sum of
scores on each dataset in the benchmark corpus: $\sum_X S_\mathtt{det}^A(X)$.

The final NAB score takes into account the detector's responsiveness to anomalies
and outputs a normalized score, \citep{lavinetal2015}, computed by
\begin{equation} \label{eq:nab_final_score}
  \mathtt{NAB\_score}_\mathtt{det}^A
    = 100 \frac{S_\mathtt{det}^A - S_\mathtt{null}^A}
               {S_\mathtt{perfect}^A - S_\mathtt{null}^A}
      \,,
\end{equation}
where $S_\mathtt{perfect}$ and $S_\mathtt{null}$ are the scores, respectively, for
the detector, which generates true positives only, and the one which outputs no alarms
at all. The range of the final score for any default profile is $(-\infty,100]$,
since the worst detector is the one which fires only false positive alarms.



\section{Benchmark Results} 
\label{sec:benchmark_results}

In this section we analyze the runtime complexity of the proposed method (sec.~%
\ref{sec:anomaly_detection_in_uts}) and conduct a comparative study on the anomaly
benchmark dataset (sec.~\ref{sec:anomaly_detection_benchmark}).

Tab.~\ref{tab:method_cplx} gives the worst case runtime complexity for the conformal
procedures in terms of the worst case complexity of the NCM $A(X_{:t}, \vec{x})$,
denoted by $c_A(t)$.
\begin{table}[t]
  \centering
  \begin{tabular}{lcc}
    \toprule
    \multirow{2}{*}{method}
      & \multicolumn{2}{c}{Prediction on series $(\vec{x}_s)_{s=1-n}^T$} \\
      & Scores &  Pv \\
    \midrule
    LDCD 
      & $T c_A(n)$
      & $T m$                    \\  
    ICAD (sliding) 
      & $T c_A(n)$
      & $T m$                    \\  
    \midrule
    ICAD (online) 
      & $T c_A(n)$
      & $T \log T$               \\
    CAD 
      & $\sum_{t=1}^T (t+n) c_A(t+n-1)$
      & $n T  + \frac12 T(T+1)$  \\
    \bottomrule
  \end{tabular}
  \caption{Worst case runtime complexity of conformal procedures on $(\vec{x}_s)_{s=1-n}^T$,
  $n$ is the length of the train sample.}
  \label{tab:method_cplx}
\end{table}
The CAD procedure is highly computationally complex: for each $\vec{x}_t$ computing
\eqref{eq:cad_scores} requires a leave-one-out-like run of $A$ over the sample of
size $t+n$ and a linear search through new non-conformity scores. In the online ICAD
it is possible to maintain a sorted array of non-conformity scores and thus compute
each $p$-value via the binary search and update the scores in one evaluation of $A$
on $\vec{x}_t$ and the \textit{reference} train sample. In the sliding ICAD and the
LDCD updating the calibration queue requires one run of $A$ as well, but computing
the $p$-value takes one full pass through $m$ scores. The key question therefore is
how severe the reliability penalty in \eqref{eq:sliding_icad_validity} is, how well
each procedure performs under non-stationarity or quasi-periodicity.

In sec.~\ref{sec:anomaly_detection_benchmark} we described a benchmark for testing
detector performance based on real-life datasets and scoring technique, which mimics
the actual costs of false negatives and false alarms. Almost all datasets in the
Numenta Benchmark and Yahoo! S5 corpora exhibit signs of quasi-periodicity or non-stationarity.
We use this benchmark to objectively measure the performance of the conformal $k$-NN
detector, proposed in sec.~\ref{sec:anomaly_detection_in_uts}.

The benchmark testing instruments provide each detector with the duration of the
``probationary'' period, which is $15\%$ of the total length of the currently used
time-series. Additionaly, the benchmark automatically calibrates each detector by
optimizing the alarm decision threshold. We use the benchmark suggested thresholds
and the probationary period duration as the size $n$ of the sliding historical window
for training and the size of the calibration queue $m$.

To measure the effect of conformal $p$-values on the performance we also test a basic
$k$-NN detector with a heuristic rule to assign confidence. Similarly to sliding
train and calibration samples in the proposed LDCD $k$-NN, the baseline $k$-NN detector
uses the train sample $\mathcal{T}_t$ as in sec.~\ref{sec:anomaly_detection_in_uts},
to compute the score of the $t$-th observation with \eqref{eq:k_nn_scorer}:
\begin{equation}
  \alpha_t = \text{NN}(\vec{x}_t; k, \mathcal{T}_t) \,.
\end{equation}
Then the score is dynamically normalized to a value within the $[0,1]$ range with
a heuristic \eqref{eq:dynrange_pv}:
\begin{equation} \label{eq:dynrange_pv}
  \mathtt{Pv}_t
    = \frac{\max_{i=0}^m \alpha_{t-i} - \alpha_t}
           {\max_{i=0}^m \alpha_{t-i} - \min_{i=0}^m \alpha_{t-i}}
    \,.
    \tag{DynR}
\end{equation}
The conformal $k$-NN detector using the LDCD procedure performs the same historical
sliding along the time-series, but its $p$-value is computed with
\eqref{eq:sliding_icad_p_value} (sec.~\ref{sec:anomaly_detection_in_uts}):
\begin{equation} \label{eq:ldcd_pv}
  \mathtt{Pv}_t
    = \frac1{m+1}
      \Bigl|\{i=0,\ldots, m\,:\,
              \alpha_{t-i} \geq \alpha_t \}\Bigr|
    \,.
    \tag{LDCD}
\end{equation}
The value $p_t = 1 - \mathtt{Pv}_t$ is the conformal abnormality score returned
by each detector for the observation $x_t$. 

We report the experiment results on two settings of $k$ and $l$ hyperparameters:
$(27, 19)$ and $(1, 1)$ for the number of neighbours $k$ and the \eqref{eq:time_delay_embed}
embedding dimension $l$ respectively. The seemingly arbitrary setting $(27, 19)$
achieved the top-3 performance in the Numenta Anomaly Detection challenge, \citep{numenta2016}.
These hyperparameter values were tuned via grid search over the accumulated performance
on the combined corpus of $\approx 400$ time series, which makes the chosen parameters
unlikely to overfit the data.

Preliminary experimental results have revealed that the \ref{eq:ldcd_pv} $k$-NN detector
has adequate anomaly coverage, but has high false positive rate. In order to decrease
the number of false alarms, we have employed the following ad hoc pruning strategy
in both detectors:
\begin{itemize}
  \item output $p_t = 1-\mathtt{Pv}_t$ for the observation $x_t$, and if $p_t$ exceeds
  $99.5\%$ fix the output at $50\%$ for the next $\frac{n}5$ observations.
\end{itemize}

The results for $k$-NN detector with $27$ neighbours and $19$-dimensional embedding 
\eqref{eq:time_delay_embed} are provided in table~\ref{tab:knn2719}.
\begin{table}[t]
  \centering
  \begin{tabular}{c>{\raggedright}p{0.2\linewidth}rrr}
    \toprule
    Corpus & $p$-value &  LowFN &  LowFP &  Standard \\
    \midrule
    \multirow{4}{*}{Numenta}
    & \ref{eq:dynrange_pv} 
                  &  -9.6 & -185.7 & -54.9 \\  
    & \ref{eq:ldcd_pv} 
                  &   4.3 & -143.8 & -34.0 \\ 
    \cmidrule{2-5}
    & \ref{eq:dynrange_pv} w. pruning
                  &  63.0 &   36.2 &  54.9 \\  
    & \ref{eq:ldcd_pv} w. pruning
                  &  64.1 &   42.6 &  56.8 \\ 
    \midrule
    \multirow{4}{*}{Yahoo!}
    & \ref{eq:dynrange_pv}
                  &   50.0 &    0.3 &   36.1 \\ 
    & \ref{eq:ldcd_pv}
                  &   50.1 &    0.4 &   36.1 \\ 
    \cmidrule{2-5}
    & \ref{eq:dynrange_pv} w. pruning
                  &   68.2 &   56.4 &   63.8 \\ 
    & \ref{eq:ldcd_pv} w. pruning
                  &   68.8 &   56.9 &   64.3 \\ 
    \bottomrule
  \end{tabular}
  \caption{NAB scores of the $k$-NN detector
  $(27, 19)$ on the Numenta and Yahoo! S5 corpora.}
  \label{tab:knn2719}
\end{table}
The key observation is that indeed the $k$-NN detector with the \ref{eq:ldcd_pv}
confidence scores performs better than the baseline \ref{eq:dynrange_pv} detector.
At the same time the abnormality score produced by the dynamic range heuristic
are not probabilistic in nature, whereas the conformal confidence scores of the
$k$-NN with the LDCD are. The rationale behind this is that conformal scores take
into account the full distribution of the calibration set, whereas the \ref{eq:dynrange_pv},
besides being simple scaling, addresses only the extreme values of the scores.

\begin{table}[b]
  \centering
  \begin{tabular}{c>{\raggedright}p{0.2\linewidth}rrr}
    \toprule
    Corpus & $p$-value &  LowFN &  LowFP &  Standard \\
    \midrule
    \multirow{4}{*}{Numenta}
    & \ref{eq:dynrange_pv}
                  & -167.0 & -658.4 & -291.0 \\ 
    & \ref{eq:ldcd_pv}
                  &   62.3 &   34.8 &   53.8 \\ 
    \cmidrule{2-5}
    & \ref{eq:dynrange_pv} w. pruning
                  &   52.2 &    4.2 &   39.0 \\ 
    & \ref{eq:ldcd_pv} w. pruning
                  &   62.7 &   30.7 &   53.5 \\ 
    \midrule
    \multirow{4}{*}{Yahoo!}
    & \ref{eq:dynrange_pv}
                  &   30.8 &  -20.7 &   16.9 \\ 
    & \ref{eq:ldcd_pv}
                  &   47.7 &   21.5 &   37.6 \\ 
    \cmidrule{2-5}
    & \ref{eq:dynrange_pv} w. pruning
                  &   50.6 &   35.2 &   44.8 \\ 
    & \ref{eq:ldcd_pv} w. pruning
                  &   53.8 &   36.2 &   46.9 \\ 
    \bottomrule
  \end{tabular}
  \caption{NAB scores of the $k$-NN detector
  $(1, 1)$ on the Numenta and Yahoo! S5 corpora.}
  \label{tab:knn0101}
\end{table}

Tab.~\ref{tab:knn0101} shows the final scores for the $k$-NN detector with $1$ neighbour
and no embedding ($l=1$). The table illustrates that the conformal LDCD procedure
works well even without alarm thinning. Heuristically, this can be explained by observing
that LDCD procedure on the $k$-NN with $1$-D embeddings in fact a sliding-window
prototype-based distribution support estimate. Furthermore, the produced $p$-values
\eqref{eq:ldcd_pv} are closely related to the probability of an extreme observation
relative to the current estimate of the support.

Tables \ref{tab:yahoo_lederboard} and \ref{tab:numenta_lederboard} show the benchmark
performance scores for detectors, which were competing in the Numenta challenge,
\citep{numenta2016}.
\begin{table}[t]
  \centering
  \begin{tabular}{>{\raggedright}p{0.35\linewidth}rrr}
    \toprule
    Detector &  LowFN &  LowFP &  Standard \\
    \midrule
    $27$-NN $l=19$ \ref{eq:ldcd_pv} w. pruning                      
        &                68.8 &                56.9 &      64.3 \\
    $1$-NN $l=1$ \ref{eq:ldcd_pv} w. pruning                        
        &                53.8 &                36.2 &      46.9 \\
    \midrule
    relativeEntropy
        &                52.5 &                40.7 &      48.0 \\
    Numenta
        &                44.4 &                37.5 &      41.0 \\
    Numenta\texttrademark
        &                42.5 &                36.6 &      39.4 \\
    bayesChangePt
        &                43.6 &                17.6 &      35.7 \\
    windowedGaussian
        &                40.7 &                25.8 &      31.1 \\
    skyline
        &                28.9 &                18.0 &      23.6 \\
    \midrule
    Random ($p_t \sim \mathcal{U}[0,1]$)
        &                47.2 &                 1.2 &      29.9 \\
    \bottomrule
  \end{tabular}
  \caption{The performance of various detectors on the Yahoo! S5 dataset.}
  \label{tab:yahoo_lederboard}
\end{table}

\begin{table}[!hb]
  \centering
  \begin{tabular}{>{\raggedright}p{0.35\linewidth}rrr}
    \toprule
    Detector &  LowFN &  LowFP &  Standard \\
    \midrule
    $27$-NN $l=19$ \ref{eq:ldcd_pv} w. pruning                      
        &                64.1 &                42.6 &      56.8 \\
    $1$-NN $l=1$ \ref{eq:ldcd_pv} w. pruning                        
        &                62.7 &                30.7 &      53.5 \\
    \midrule
    Numenta
        &                74.3 &                63.1 &      70.1 \\
    Numenta\texttrademark
        &                69.2 &                56.7 &      64.6 \\
    relativeEntropy
        &                58.8 &                47.6 &      54.6 \\
    windowedGaussian
        &                47.4 &                20.9 &      39.6 \\
    skyline
        &                44.5 &                27.1 &      35.7 \\
    bayesChangePt
        &                32.3 &                 3.2 &      17.7 \\
    \midrule
    Random ($p_t \sim \mathcal{U}[0,1]$)
        &                25.9 &                 5.8 &      16.8 \\
  \bottomrule
  \end{tabular}
  \caption{The performance of various detectors on the Numenta dataset.}
  \label{tab:numenta_lederboard}
\end{table}



\section{Conclusion} 
\label{sec:conclusion}

In this paper we proposed a conformal $k$-NN anomaly detector for univariate time
series, which uses sliding historical windows both to embed the time series into
a higher dimensional space for $k$-NN and to keep the most relevant observations
to explicitly address potential quasi-periodicity. The proposed detector was tested
using a stringent benchmarking procedure \citep{lavinetal2015}, which mimics the
real costs of timely signals, tardy alarms and misdetections. Furthermore we supplemented
the benchmark dataset corpus with Yahoo! S5 anomaly dataset to cover more use-cases.
The results obtained in sec.~\ref{sec:benchmark_results} demonstrate that the conformal
$k$-NN has adequate anomaly coverage rate and low false negative score. The cases,
when the conformal LDCD scores required the use of a signal pruning step, were also
the cases when the baseline $k$-NN detector was over-sensitive. Nevertheless, in all
cases, conformal abnormality confidence scores improved the benchmark scores.

Numenta held a detector competition in 2016 in which the prototype of the proposed
procedure, \citep{2016arXiv160804585B}, took the third place, \citep{numenta2016},
competing against much more complex methods based on cortical memory, neural networks,
etc. The favourable results on the NAB corpus (sec.~\ref{sec:benchmark_results})
suggest that the theoretical foundations of the LDCD procedure, specifically the
assumptions required for the proper validity guarantee, should be subject of further
research. Besides the validity guarantees, the effects of the violations of the iid
assumption should be investigated as well, especially since the embedded time-series
vectors overlap.


\bibliography{references/references.bib}

\end{document}